%% file: Main.tex
\def\year{2022}\relax
\documentclass[letterpaper]{article} % DO NOT CHANGE THIS
\usepackage{aaai22}  % DO NOT CHANGE THIS
\usepackage{times}  % DO NOT CHANGE THIS
\usepackage{helvet}  % DO NOT CHANGE THIS
\usepackage{courier}  % DO NOT CHANGE THIS
\usepackage[hyphens]{url}  % DO NOT CHANGE THIS
\usepackage{graphicx} % DO NOT CHANGE THIS
\def\UrlFont{\rm}  % DO NOT CHANGE THIS
\usepackage{natbib}  % DO NOT CHANGE THIS AND DO NOT ADD ANY OPTIONS TO IT
\usepackage{caption} % DO NOT CHANGE THIS AND DO NOT ADD ANY OPTIONS TO IT
\DeclareCaptionStyle{ruled}{labelfont=normalfont,labelsep=colon,strut=off} % DO NOT CHANGE THIS
\usepackage{algorithm}
\usepackage{algorithmic}

\usepackage{multirow}
\usepackage{float} 
\usepackage{subfigure}

\usepackage{xcolor}
\usepackage{amsmath}
\usepackage{amssymb}
\usepackage{booktabs}
\usepackage{newfloat}
\usepackage{listings}
\floatstyle{ruled}
\newfloat{listing}{tb}{lst}{}
\floatname{listing}{Listing}
\title{Evo-ViT: Slow-Fast Token Evolution for Dynamic Vision Transformer}
\author{Paper ID: 1350}
\title{Evo-ViT: Slow-Fast Token Evolution for Dynamic Vision Transformer}
\author{Yifan Xu\textsuperscript{\rm 1,3,4}\thanks{The first two authors contributed equally. 
}
, Zhijie Zhang\textsuperscript{\rm 2,3$*$}, Mengdan Zhang\textsuperscript{\rm 3}, Kekai Sheng\textsuperscript{\rm 3}, Ke Li\textsuperscript{\rm 3}, Weiming Dong\textsuperscript{\rm 1,4}\thanks{The co-corresponding authors.}
, \\Liqing Zhang\textsuperscript{\rm 2}, Changsheng Xu\textsuperscript{\rm 1,4}, Xing Sun\textsuperscript{\rm 3$\dagger$}\\}
\affiliations{
\textsuperscript{\rm 1}Institute  of  Automation,  Chinese  Academy  of Sciences  
\textsuperscript{\rm 2}Shanghai Jiao Tong University  
\textsuperscript{\rm 3}Tencent Youtu Lab\\
\textsuperscript{\rm 4}School of Artificial Intelligence, University of Chinese Academy of Sciences\\
\{xuyifan2019, weiming.dong, changsheng.xu\}@ia.ac.cn, zzj506506@sjtu.edu.cn, zhang-lq@cs.sjtu.edu.cn,\\
\{davinazhang, saulsheng, tristanli\}@tencent.com, winfred.sun@gmail.com
}
\begin{document}
\maketitle
\input{Sections/Abstract}

\input{Sections/Introduction_v2}

\input{Sections/Related_Works_v2}

\input{Sections/Background}

\input{Sections/Methodology_v2}

\input{Sections/Experiments_v3}

\input{Sections/Conclusions}

\section{Acknowledgements}
This work was supported by National Key R\&D Program of China under no. 2020AAA0106200, by Shanghai Municipal Science and Technology Key Project under no. 20511100300, by National Natural Science Foundation of China under nos. 62076162, U20B2070, 61832016 and 62036012, and by CASIA-Tencent Youtu joint research project.

% \small{
\bibliography{Main}
% }

\end{document}

%% file: Sections/Abstract.tex
\begin{abstract}

Vision transformers (ViTs) have recently received explosive popularity, but the huge computational cost is still a severe issue. Since the computation complexity of ViT is quadratic with respect to the input sequence length, a mainstream paradigm for computation reduction is to reduce the number of tokens. Existing designs include structured spatial compression that uses a progressive shrinking pyramid to reduce the computations of large feature maps, and unstructured token pruning that dynamically drops redundant tokens. However, the limitation of existing token pruning lies in two folds: 
1) the incomplete spatial structure caused by pruning is not compatible with structured spatial compression that is commonly used in modern deep-narrow transformers;
% 2) it is usually conducted after a time-consuming pre-training. 
2) it usually requires a time-consuming pre-training procedure.
To tackle the limitations and expand the applicable scenario of token pruning, we present Evo-ViT, a self-motivated slow-fast token  evolution approach for vision transformers. Specifically, we conduct unstructured instance-wise token selection by taking advantage of the  simple and effective global class attention that is native to vision transformers. Then, we propose to update the selected informative tokens and uninformative tokens with different computation paths, namely, slow-fast updating. Since slow-fast updating mechanism maintains the spatial structure and information flow, Evo-ViT can accelerate vanilla transformers of both flat and deep-narrow structures from the very beginning of the training process. Experimental results demonstrate that our method significantly reduces the computational cost of vision transformers while maintaining comparable performance on image classification. For example, our method accelerates DeiT-S by over 60$\%$ throughput while only sacrificing 0.4$\%$ top-1 accuracy on ImageNet-1K, outperforming current token pruning methods on both accuracy and efficiency.
Code is available at \textcolor{gray}{\url{https://github.com/YifanXu74/Evo-ViT}}.
\end{abstract}

%% file: Sections/Introduction_v2.tex
\section{Introduction}
\label{sec:introduction}

%Recently, transformers, dominated in natural language processing (NLP), have show strong power on various computer vision tasks such as image classification, object detection, instance segmentation and low-level generalization. The reason of introducing transformers into computer vision lies on its unique properties that convolution neural networks (CNN) lack, especially the property of modeling long-range dependencies in the data. However, different from NLP tasks, where most tokens are important due to the explicit grammatical structure, the modeling of long-range dependencies among image tokens across layers of the transformer usually brings computation inefficiency because the image may contain a large region of low frequency context or uninformative backgrounds. 
%%%%davina
Recently, vision transformers (ViTs) have shown strong power on various computer vision tasks~\cite{survey4}.
% such as image classification~\cite{CeiT}, object detection~\cite{DETR}, and instance segmentation~\cite{SETR}. 
The reason of introducing the transformer into computer vision lies in its unique properties that convolution neural networks (CNNs) lack, especially the property of modeling long-range dependencies. However, dense modeling of long-range dependencies among image tokens brings computation inefficiency, because images contain large regions of low-level texture and uninformative background.

%To address the inefficiency problem of modeling long-range dependencies in vision transformers, existing methods follow three mainstreams. The first is to improve the computation efficiency of the multi-head self-attention (MSA) module~\cite{pvt, cvt, swin_transformer}. They structurally reducing the number of key and value embeddings with local-wise linear projection or convolutional projection.  Although achieving locality prior and efficiency, this kind of  sub-sampling takes object and background equally, which is not discriminative enough. The second paradigm is network pruning. (cite Patch slimming method) develops a top-down patch slimming approach that can identify and remove redundant patches in the pre-defined network. The whole pruning scheme is conducted layer-wisely under a careful control of the network error, which is of high accuracy but not flexible enough for network training. The pruning mask is fixed for all instances, which seems running counter to the property of dynamic relation modeling of vision transformers. The third paradigm is token sparsification. DynamicViT explores an unstructured and data-dependent downsampling strategy based on pre-trained vision transformers. A binary decision mask is learnt by conducting linear projections on token representations to abandon tokens of less importance in the inference stage. However, this kind of unstructured sparsification method is hard to be applied to vision transformers of the deep-narrow structure, which affects its extensibility to downstream tasks.

\input{Figs/tech_pipline}

Existing methods follow two pipelines to address the inefficiency problem of modeling long-range dependencies among tokens in ViT as shown in the above two pathways of Fig.~\ref{fig:tech_pipline}. The first pipeline, as shown in the second pathway, is to perform structured compression based on local spatial prior,
such as local linear projection~\cite{PVT}, convolutional projection~\cite{PiT}, and shift windows~\cite{SwinT}. Most modern transformers with deep-narrow structures are within this pipeline. However, the structured compressed models treat the informative object tokens and uninformative background tokens with the same priority.
Thus, token pruning, the second pipeline, proposes to identify and drop the uninformative tokens in an unstructured way. \cite{PatchSlimming} improves the efficiency of a pre-trained transformer network by developing a top-down layer-by-layer token slimming approach that can identify and remove redundant tokens based on the reconstruction error of the pre-trained network. The trained pruning mask is fixed for all instances. \cite{DynamicViT} proposes to accelerate a pre-trained transformer network by removing redundant tokens hierarchically, and explores an data-dependent down-sampling strategy via self-distillation.
Despite of the significant acceleration, these unstructured token pruning methods are restricted in two folds due to their incomplete spatial structure and information flow, namely, the inapplicability on structured compressed transformers and inability to train from scratch.

In this paper, as shown in the third pathway of Fig.~\ref{fig:tech_pipline}, we propose to handle the inefficiency problem in a dynamic data-dependent way from the very beginning of the training process while suitable for structured compression methods. We denote uninformative tokens that contribute little to the final prediction but bring computational cost when bridging redundant long-range dependencies as placeholder tokens. Different from structured compression that reduces local spatial redundancy in ~\cite{PVT, LeViT}, we propose to distinguish the informative tokens from the placeholder tokens for each instance in an unstructured and dynamic way, and update the two types of tokens with different computation paths. 
Instead of searching for redundancy and pruning in a pre-trained network such as \cite{PatchSlimming, DynamicViT}, by preserving placeholder tokens, 
the complete spatial structure and information flow can be maintained.
% the redundancy problem can be alleviated in the beginning of the training process of a new model,
In this way, our method can be a generic plugin in most ViTs of both flat and deep-narrow structures from the very beginning of training.

%The common observation is that for a pre-trained transformer, there indeed exists uninformative tokens that contribute little to the final prediction but bring computational cost when bridging redundant long-range dependencies, especially in deeper layers. We denote these tokens as Placeholder Tokens. In contrast to searching for redundancy in a pre-trained network, benefiting from structural compression that compresses a flat transformer like ViT~\cite{ViT} to  a pyramid-like transformer~\cite{PVT, CVT, SwinT}, the redundancy problem can be alleviated in the beginning of the training process of a new model. However, structural compression usually handles spatial redundancy and does not specifically distinguish discriminative tokens from placeholder tokens for each instance. Thus, its ability of removing redundancy is limited.  

%In this paper, we propose to handle the redundancy problem from the very beginning of the training process of a versatile transformer and improve its training and inference efficiency. It is achieved by specifically identifying  discriminative tokens and placeholer tokens for each instance in a transformer, and updating them in a slow-fast way. The benefits of preserving the placeholer tokens lie in two folds: first, it ensures complete information flow in a transformer for modeling accuracy; second, the structure preserving design is friendly for applications on existing sota transformers of both flat and pyramid-like (deep-narrow) structures, and is also friendly for downstream tasks.
\input{Figs/fig1}

Concretely, 
% Evo-ViT~\footnote{Code is available at \url{https://github.com/YifanXu74/Evo-ViT}.}
% \iffalse
% \footnote{The code is available at https://github.com/YifanXu74/Evo-ViT.}
% \fi
Evo-ViT, a self-motivated slow-fast token evolution approach for dynamic ViTs is proposed in this work. ``Self-motivated'' means that transformers can naturally distinguish informative tokens from placeholder tokens for each instance, since they have insights into global dependencies among image tokens. 
Without loss of generality, we take DeiT~\cite{DeiT} in Fig.~\ref{fig:fig1} as example. We find that the class token of DeiT-T estimates importance of each token for dependency modeling and final classification. Especially in deeper layers (\emph{e.g.}, layer 10), the class token usually augments informative tokens with higher attention scores and has a sparse attention response, which is quite consistent to the visualization result provided by~\cite{Trans_interpretability} for transformer interpretability. In shallow layers (\emph{e.g.}, layer 5), the attention of the class token is relatively scattered but mainly focused on informative regions. Thus, taking advantage of class tokens, informative tokens and placeholer tokens are determined. The preserved placeholer tokens ensure complete information flow in shallow layers of a transformer for modeling accuracy. 
After the two kinds of tokens are determined, they are updated in a slow-fast approach. Specifically, the placeholder tokens are summarized to a representative token that is evolved via the full transformer encoder simultaneously with the informative tokens in a slow and elaborate way. Then, the evolved representative token is exploited to fast update the placeholder tokens for more representative features.

We evaluate the effectiveness of the proposed Evo-ViT method on two kinds of baseline models, namely, transformers of flat structures such as DeiT~\cite{DeiT} and transformers of deep-narrow structures such as LeViT~\cite{LeViT} on ImageNet~\cite{imagenet} dataset.  Our self-motivated slow-fast token evolution method allows
the DeiT model to improve inference throughput by 40\%-60\% and further accelerates the state-of-the-art efficient transformer LeViT while maintaining comparable performance. 

%Results  demonstrate that by pruning 50\% tokens from the forth layers of DeiT leads to over 50\% improvement on throughput with almost no accuracy drop.

%% file: Figs/tech_pipline.tex
\begin{figure}[t]
    \centering
    \includegraphics[width=0.45\textwidth]{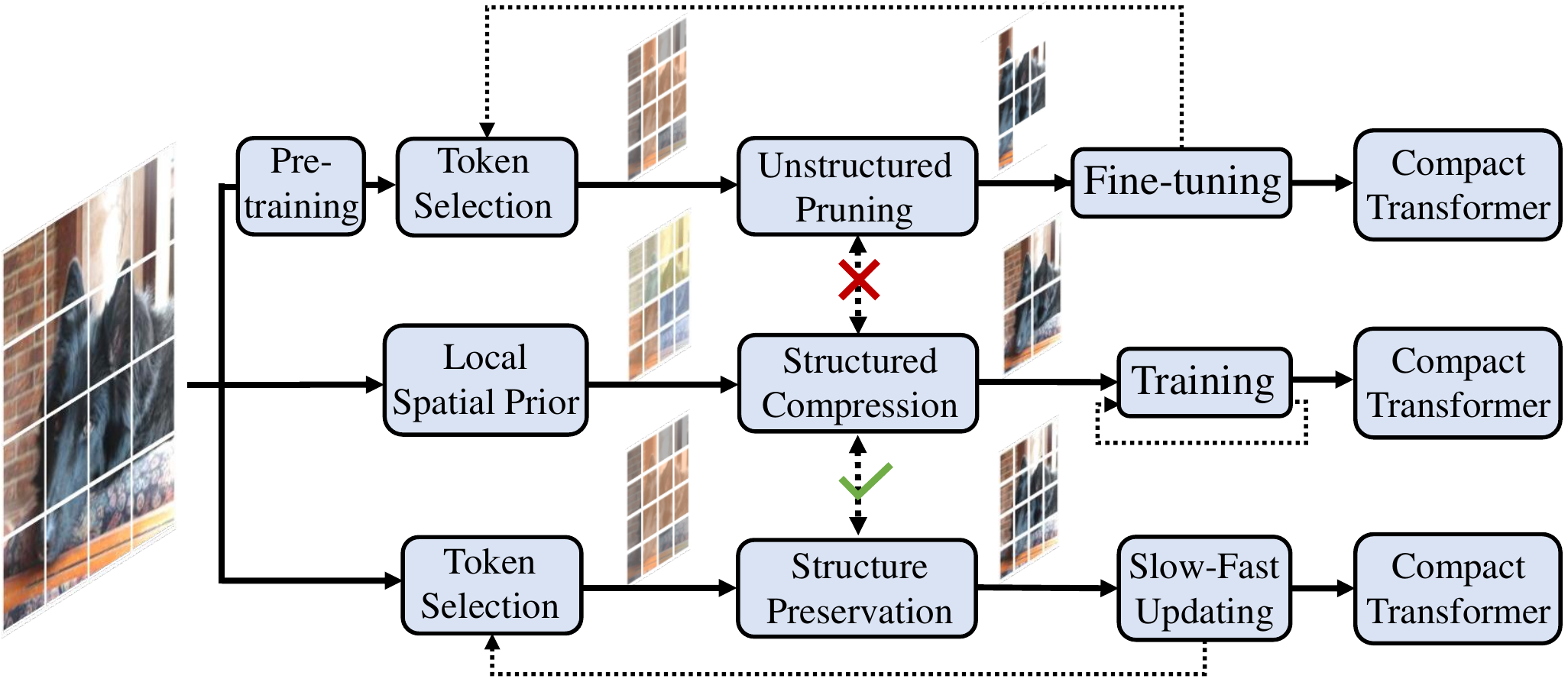} % Reduce the figure size so that it is slightly narrower than the column.
    \caption{An illustration of technique pipelines for computation reduction via tokens. The dash lines denote iterative training. The first branch: the pipeline of unstructured token pruning~\cite{DynamicViT,PatchSlimming} based on pre-trained models. The second branch: the pipeline of structured compression~\cite{LeViT}. The third branch: our proposed pipeline that performs unstructured updating while suitable for structured compressed models.}
    \label{fig:tech_pipline}
    \vspace{-4mm}
\end{figure}

%% file: Figs/fig1.tex
\begin{figure}[t]
    \centering
    \includegraphics[width=0.45\textwidth]{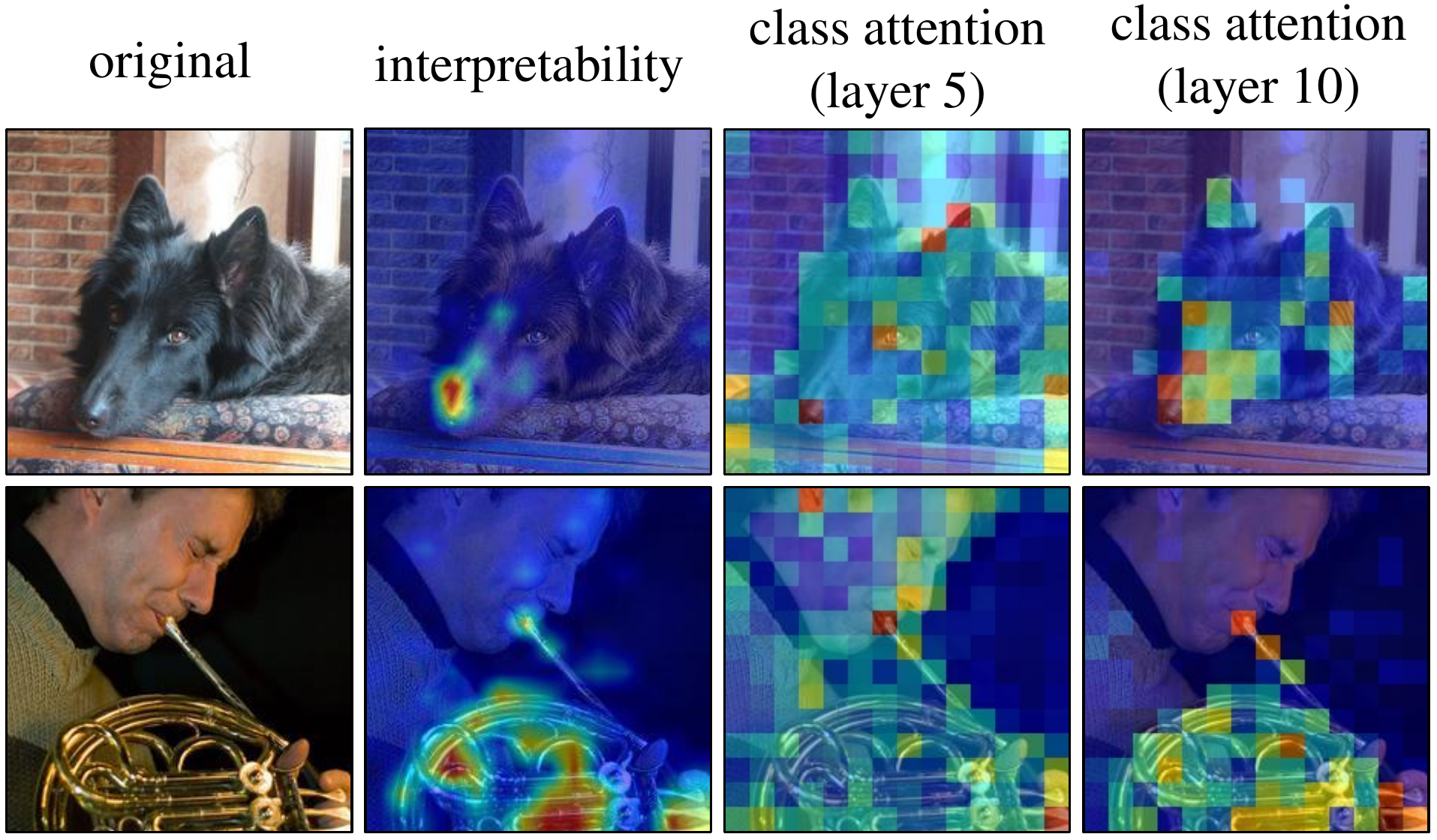} % Reduce the figure size so that it is slightly narrower than the column.
    \caption{Visualization of class attention in DeiT-T. The \textit{interpretability} comes from ~\cite{Trans_interpretability}.}
    \label{fig:fig1}
    \vspace{-5mm}
\end{figure}

%% file: Sections/Related_Works_v2.tex
\section{Related Work}
\paragraph{Vision Transformer.}
Recently, a series of transformer models~\cite{survey1,survey4} are proposed to solve various computer vision tasks. The transformer has achieved promising success in image classification~\cite{ViT,DeiT,ConViT}, object detection~\cite{DETR,SwinT,deformableDETR} and instance segmentation~\cite{sstvos,SETR} due to its significant capability of modeling long-range dependencies. 
Vision Transformer (ViT)~\cite{ViT} is among the pioneering works that achieve state-of-the-art performance with large-scale pre-training. DeiT~\cite{DeiT} manages to tackle the data-inefficiency problem in ViT by simply adjusting training strategies and adding an additional token along with the class token for knowledge distillation. 
To achieve better accuracy-speed trade-offs for general dense prediction, recent works~\cite{T2T,LeViT,PVT} design transformers of deep-narrow structures by adopting sub-sampling operation (\emph{e.g.}, strided down-sampling, local average pooling, convolutional sampling) to reduce the number of tokens in intermediate layers.

\paragraph{Redundancy Reduction.}
Transformers take high computational cost because the multi-head self-attention (MSA) requires quadratic time complexity and the feed forward network (FFN) increases the dimension of latent features. 
The existing acceleration methods for transformers can be mainly categorized into sparse attention mechanism, knowledge distillation~\cite{distilbert}, and pruning.  The sparse attention mechanism includes, for example, low rank factorization~\cite{CoaT,Linformer}, fixed local patterns~\cite{SwinT}, and learnable patterns~\cite{Synthesizer,longformer}. \cite{PS-ViT-ICCV} handles the inefficiency problem by sparse input patch sampling.
% More specifically, \cite{SwinT} proposes a shifted windowing scheme that brings greater efficiency by limiting self-attention computation to non-overlapping local windows while also allowing for cross-window connection.  
\cite{Realformer} proposes to add an evolved global attention to the attention matrix in each layer for a better residual mechanism. Motivated by this work, we propose the evolved global class attention to guide the token selection in each layer.
% The Linformer \cite{Linformer} is a classic example of low rank methods as it projects the length dimension of keys and values to a lower-dimensional representation. 
% Realformer
The closest paradigm to this work is token pruning. \cite{PatchSlimming} presents a top-down layer-by-layer patch slimming algorithm to reduce the computational cost in pre-trained vision transformers. The patch slimming scheme is conducted under a careful control of the feature reconstruction error, so that the pruned transformer network can maintain the original performance with lower computational cost. \cite{DynamicViT} devises a lightweight prediction module to estimate the importance score of each token given the current features of a pre-trained transformer. The module is plugged into different layers to prune placeholder tokens in a unstructured way and is supervised by a distillation loss based on the predictions of the original pre-trained transformer. Different from these pruning works, we proposed to preserve the  placeholder tokens, and update the informative tokens and placeholder tokens with different computation paths; thus our method can achieve better performance and be suitable for various transformers due to the complete spatial structure. 
In addition, the complete information flow allows us to accelerate transformers  with scratch training. 

%% file: Sections/Background.tex
\section{Preliminaries}
\label{sec:background}
% class token attention
% Preliminaries: Vision Transformers

ViT~\cite{ViT} proposes a simple tokenization strategy that handles images by reshaping them into flattened sequential patches and linearly projecting each patch into latent embedding. An extra class token (CLS) is added to the sequence and serves as the global image representation. Moreover, since self-attention in the transformer encoder is position-agnostic and vision applications highly require position information, ViT adds position embedding into each token, including the CLS token. Afterwards, all tokens are passed through stacked transformer encoders and the CLS token is used for final classification.

The transformer is composed of a series of stacked encoders where each encoder consists of two modules, namely, a multi-head self-attention (MSA) module and a feed forward network (FFN) module. The FFN module  contains two linear transformations with an activation function. The residual connections are employed around both MSA and FFN modules, followed by layer normalization (LN).  Given the input $x_{0}$ of ViT, the processing of the k-th encoder can be mathematically expressed as
\begin{equation}
    \begin{split}
        x_{0} &= \lceil x_{cls} |\ x_{patch} \rceil + x_{pos}, \\
        y_{k} &= x_{k-1} + MSA(LN(x_{k-1})), \\
        x_{k} &= y_{k} + FFN(LN(y_{k})),\\
    \end{split}
\end{equation}
where $x_{cls}\in \mathbb{R}^{1 \times C}$ and $x_{patch}\in \mathbb{R}^{N \times C}$ are CLS and patch tokens respectively and $x_{pos}\in \mathbb{R}^{(1+N) \times C}$ denotes the position embedding. $N$ and $C$ are the number of patch tokens and the dimension of the embedding.

%The vanilla vision transformer \cite{} lays the foundation of our framework, which consists of many transformer blocks. Each transformer block includes a Multi-head self-attention(MSA) and a feed forward network(FFN). 
%Particularly, we follow the tokenized strategy to handle 2D image and reshape the original input into flattened sequential patches $X\in \mathbb{R}^{N \times C}$, where $N$ indicates the number of tokens while $C$ is latent dimension size. 
Specifically, a self-attention (SA) module projects the input sequences into  query, key, value vectors (\emph{i.e.}, $Q, K, V \in \mathbb{R}^{(1+N) \times C}$) using three learnable linear mapping $W_Q$, $W_K$ and $W_V$.  Then, a weighted sum over all values in the sequence is computed through:
%scaled dot-product with a softmax function to calculate similarity scores between each two tokens. The outputs are computed by weighted sum of similarity scores and values. The multi-head self-attention can be formulated as:
\begin{equation}
   SA(Q,K,V)=Softmax(\cfrac{Q \cdot K^T}{\sqrt{C}}) V.
%    head_i=SA(Q_i,K_i,V_i)\\
%    MSA(Q,K,V)=Concat(head_1,head_2, \cdots, head_H)
\end{equation}
%where H is the number of heads. 
MSA is an extension of SA. It splits queries, keys, and values for $h$ times and performs the attention function in parallel, then linearly projects their concatenated outputs. 

It is worth noting that one very different design of ViT from CNNs is the CLS token. The CLS token interacts with patch tokens at each encoder and summarizes all the patch tokens for the final representation. We denote the similarity scores between the CLS token and patch tokens as class attention $A_{cls}$, formulated as:
\begin{equation}
    A_{cls} = Softmax(\cfrac{q_{cls} \cdot K^T}{\sqrt{C}}),
    \label{eq:cls_attn}
\end{equation}
where $q_{cls}$ is the query vector of the CLS token. 

\noindent\textbf{Computational complexity.}
In ViT, the computational cost of the MSA and FFN modules are $O(4NC^{2}+2N^{2}C)$ and $O(8NC^{2})$, respectively. For pruning methods~\cite{DynamicViT, PatchSlimming}, by pruning $\eta\%$ tokens, at least $\eta\%$ FLOPS in the FFN and MSA modules can be reduced. Our method can achieve the same efficiency while suitable for scratch training and versatile downstream applications.

%% file: Sections/Methodology_v2.tex
\section{Methodology}
\label{sec:methodlogy}
\subsection{Overview}

\input{Figs/fig_method}
In this paper, we aim to handle the inefficiency modeling issue in each input instance from the very beginning of the training process of a versatile transformer. 
As shown in Fig~\ref{fig:method}, the pipeline of Evo-ViT mainly contains two parts: the structure preserving token selection module and the slow-fast token updating module. In the structure preserving token selection module, the informative tokens and the placeholder tokens are determined by the evolved global class attention, so that they can be updated in different manners in the following slow-fast token updating module. Specifically, the placeholder tokens are summarized and updated by a representative token. The long-term dependencies and feature richness of the representative token and the informative tokens are evolved via the MSA and FFN modules.  

We first elaborate on the proposed structure preserving token selection module. Then, we introduce how to update the informative tokens and the placeholder tokens in a slow-fast approach. Finally, the training details, such as the loss and other training strategies, are introduced.

\input{Figs/fig_cka}

% \vspace{-3mm}
\subsection{Structure preserving token selection}
\label{sec:cls_attn_analysis}
% 基于class token进行选择
% global class token
In this work, we propose to preserve all the tokens and dynamically distinguish informative tokens and placeholder tokens for complete information flow. The reason is that it is  not trivial to prune tokens in shallow and middle layers of a vision transformer, especially in the beginning of the training process. We explain this problem in both inter-layer and intra-layer ways. 
First, shallow and middle layers usually present fast growing capability of feature representation. Pruning tokens brings severe information loss. 
Following Refiner~\cite{Refiner}, we use centered kernel alignment (CKA) similarity~\cite{CKA} to measure similarity of the intermediate token features in each layer and the final CLS token, assuming that the final CLS token is strongly correlated with classification.  As shown in Fig.~\ref{fig:cka_score}, the token features of DeiT-T keep evolving fast when the model goes deeper and the final CLS token feature is quite different from token features in shallow layers. It indicates that the representations in shallow or middle layers are insufficiently encoded, which makes token pruning quite difficult. 
Second, tokens have low correlation with each other in the shallow layers.
We evaluate the Pearson correlation coefficient (PCC) among different patch token queries with respect to the network depth in the DeiT-S model to show 
redundancy. As shown in Fig.~\ref{fig:token_correlation}, the lower correlation with larger variance in the shallow layers also proves the difficulty to distinguish redundancy in shallow features. 
% An interesting finding is that the first two layers have higher correlation and smaller variance among tokens, indicating the potential relationship between local patterns and token correlation.

The attention weight is the easiest and most popular approach~\cite{Attention_rollout,EvolvingAttn} to interpret a model’s decisions and to gain insights about the propagation of information among tokens. The class attention weight described in Eqn.~\ref{eq:cls_attn} reflects the information collection and broadcast processes of the CLS token. We find that our proposed evolved global class attention is able to be a simple measure to help dynamically distinguish informative tokens and placeholder tokens in a vision transformer. In Fig.~\ref{fig:cka_score}, the distinguished informative tokens have high CKA correlation with the final CLS token, while the placeholder tokens have low CKA correlation.
As shown in Fig.~\ref{fig:fig1},  the global class attention is able to focus on the object tokens, which is consistent to the visualization results of~\cite{Trans_interpretability}. In the following part of this section, detailed introduction of our structure preserving token selection method is provided.

%Since the effectiveness of class token to redundancy recognition, a natural idea is to utilize class token attention to guide informative token selection. 
%As shown in Figure~\ref{fig:method}, we propose a token selection paradigm based on global class token attention.
%For easier understanding, we first introduce a naive token selection paradigm based on the original class token attention and our full method replaces the class token attention with a global class token attention for better stability. 

%Given $N$ input patch tokens $x \in \mathbb{R}^{N \times C}$ and their corresponding class token attention $A = \{A_{i}\}_{1}^{N}$, we select $k$ tokens whose class token attention is among the top $k$ of $A$ as the informative tokens. The remaining $N-k$ tokens are recognized as placeholder tokens that contain less information. Note that the placeholder tokens are kept and fast-updated rather than directly dropped.

As discussed in Preliminaries Section, the class attention $A_{cls}$  is calculated by Eqn.~\ref{eq:cls_attn}. We select $k$ tokens whose scores in the class attention are among the top $k$ as the informative tokens. The remaining $N-k$ tokens are recognized as placeholder tokens that contain less information. Different from token pruning, the placeholder tokens are kept and fast-updated rather than dropped.

For better capability of capturing the underlying information among tokens in different layers, we propose a global class attention that augments the class attention by evolving it across layers as shown in Fig.~\ref{fig:method}. Specifically, a residual connection between class attention of different layers is designed to facilitate the attention information flow with some regularization effects. Mathematically, 
\begin{equation}
    A_{cls, g}^{k} = \alpha \cdot A_{cls, g}^{k-1} + (1-\alpha)\cdot A_{cls}^{k},
    \label{eq:global_attn}
\end{equation}
where $A_{cls, g}^{k}$ is the global class attention in the k-th layer, and $A_{cls}^{k}$ is the class attention in the k-th layer. We use $A_{cls, g}^{k}$ for the token selection in the (k+1)-th layer for stability and efficiency. For each layer with token selection, only the global class attention scores of the selected informative tokens are updated.
%Now the last question is how to get the class token attention $A$. A natural idea is to calculate the class attention score of each token at the beginning of each xxx block. However, this leads into nonnegligible additional computational cost, which is inconsistent with the acceleration goal. Thus, we propose a global class attention as the selection metrics. As shown in figure x, the global class attention is initialized by the first class token attention. In each self-attention module, we only need to update the global score of selected informative tokens by adding their class attention scores to corresponding global ones. During the token selection, the informative tokens are selected by current global class attention, which avoids re-calculating the class attention. Equation xx illustrate the mathematical explanation. 

% \vspace{-3mm}
\subsection{Slow-fast token updating}
% Slow update
% Fast update
% 为什么work？一些分析还是欠缺
Once the informative tokens and the placeholder tokens are determined by the global class attention, we propose to update tokens in a slow-fast way instead of harshly dropping placeholder tokens as \cite{PatchSlimming,DynamicViT}. As shown in Fig.~\ref{fig:method}, informative tokens are carefully evolved via MSA and FFN modules, while placeholder tokens are coarsely summarized and updated via a representative token. 
%Fig.~\ref{fig:swinsulit} further illustrates the necessity of preserving placeholder tokens: swimsuits are always of higher probability than dresses with ocean backgrounds (need modified, ambiguous). Thus, though placeholder tokens are usually less informative and bring redundant relationship modeling, they are sometimes beneficial to the final results due to their causal relationship to informative tokens of main objects. 
We introduce our slow-fast token updating strategy mathematically as follows.

%During token selection, we denote the unselected tokens with less information as placeholder tokens.  As shown in figure x, the placeholder tokens usually represent background. Though of less importance, placeholder tokens are sometimes beneficial to the final results due to their causal relationship to informative tokens of main objects. For example, swimsuits (need substituted) are always of higher probability than dresses with ocean backgrounds. Thus, instead of being simply kept, placeholder tokens can bring more power with some fast and coarse updating. In our xxx method, we propose a fast-slow update strategy to address informative tokens and placeholder tokens differently.

% slow update
For $N$ patch tokens $x_{patch}$, we first split them into $k$ informative tokens $x_{inf} \in \mathbb{R}^{k \times C}$ and $N-k$ placeholder tokens $x_{ph} \in \mathbb{R}^{(N-k) \times C}$ by the above-mentioned token selection strategy. Then, the placeholder tokens $x_{ph}$ are aggregated into  a representative token $x_{rep} \in \mathbb{R}^{1 \times C}$ , as follows:
\begin{equation}
    x_{rep}=\phi_{agg}(x_{ph}), 
    \label{eq:aggregation}
\end{equation}
where $\phi_{agg}: \mathbb{R}^{(N-k) \times C} \rightarrow \mathbb{R}^{1 \times C}$ denotes an aggregating function, such as weighted sum or transposed linear projection~\cite{mlp-mixer}. Here we use weighted sum based on the corresponding global attention score in Eqn.~\ref{eq:global_attn}.

%The informative tokens are sent to pass a slow and elaborate updating, namely the full transformer modules including MSA and FFN. 
%For the placeholder tokens, we propose a representative token for fast updating. Firstly, the placeholder tokens $x_{inf}$ are aggregated into  a representative tokens $x_{rep} \in \mathbb{R}^{1 \times C}$ , as:
%\begin{equation}
%    x_{rep}=\phi_{agg}(x_{ph}), 
%    \label{eq:aggregation}
%\end{equation}
%where $\phi_{agg}: \mathbb{R}^{(N-k) \times C} \rightarrow \mathbb{R}^{1 \times C}$ denotes an aggregating function such as weighted pooling or transposed linear functions. Here we use weighted pooling based on the corresponding global attention score in Eqn.~\ref{eq:global_attn} for convenience. 

Then, both the informative tokens $x_{inf}$ and the representative token $x_{rep}$ are fed into MSA and FFN modules, and their residuals are recorded as $x_{inf}^{(*)}$ and $x_{rep}^{(*)}$ for skip connections, which can be denoted by:
\begin{equation}
    \begin{split}
        x_{inf}^{(1)}, \ x_{rep}^{(1)} & = MSA(x_{inf},x_{rep}),\\
        x_{inf} \gets & x_{inf} + x_{inf}^{(1)}, \
        x_{rep} \gets x_{rep} + x_{rep}^{(1)},\\
        x_{inf}^{(2)}, \ x_{rep}^{(2)} & = FFN(x_{inf},x_{rep}),\\
        x_{inf} & \gets x_{inf} + x_{inf}^{(2)}.\\
    \end{split}
\end{equation}
Thus, the informative tokens $x_{inf}$ and the representative token $x_{rep}$ are updated in a slow and elaborate way.

Finally, the placeholder tokens $x_{ph}$ are updated in a fast way by the residuals of $x_{rep}$:
\begin{equation}
    x_{ph} \gets x_{ph} + \phi_{exp}(x_{rep}^{(1)}) + \phi_{exp}(x_{rep}^{(2)}),
\end{equation}
where $\phi_{exp}: \mathbb{R}^{1 \times C} \rightarrow \mathbb{R}^{(N-k) \times C}$ denotes an expanding function, such as simple copy in our method. 

% why use residual
It is worth noting that the placeholder tokens are fast updated by the residuals of $x_{rep}$ rather than the output features. In fact, the fast updating serves as a skip connection for the placeholder tokens.
By utilizing residuals, we can ensure the output features of the slow updating and fast updating modules within the same order of magnitude. 
% Simply utilizing the output features for fast updating will lead to inconsistent order of magnitude among the output features of the slow updating and fast updating modules

% \vspace{-3mm}
\subsection{Training Strategies}
% \subsection{Training and inference}
% layer-stage：稳定加速
% Avg pool + cls token 训练时加监督
% 强调train from scratch

% 为什么能train from scratch?

\noindent\textbf{Layer-to-stage training schedule.} 
Our proposed token selection mechanism becomes increasingly stable and consistent as the training process. Fig.~\ref{fig:prune_vis} shows that the token selection results of a well-trained transformer turn to be consistent across different layers;  thereby indicating that the transformer tends to augment informative tokens with computing resource as much as possible, namely the full transformer networks.
Thus, we propose a layer-to-stage training strategy for further efficiency. Specifically, we conduct the token selection and slow-fast token updating layer by layer at the first 200 training epochs. During the remaining 100 epochs, we only conduct token selection at the beginning of each stage, and then slow-fast updating is normally performed in each layer. For transformers with flat structure such as DeiT, we manually arrange four layers as one stage. 
%Namely, we select the informative tokens and fast update the placeholder tokens layer by layer at the first 200 epochs. During the remaining 100 epochs, we only conduct token selection at the beginning of each stage and update tokens consistently among each stage.

\noindent\textbf{Assisted CLS token loss.}
Although many state-of-the-art vision transformers~\cite{PVT,LeViT} remove the CLS token and use the final average pooled features for classification, it is not difficult to add a CLS token in their models for our token selection strategy. We empirically find that the ability of distinguishing two types of tokens of the CLS token as illustrated in Fig.~\ref{fig:fig1} is kept in these models even without supervision on the CLS token. For better stability, we calculate classification losses based on the CLS token together with the final average pooled features during training. Mathematically, 
\begin{equation}
    \begin{split}
        \hat{y}_{cls},\hat{y}&=m(x_{cls},x_{patch}), \\
        \mathcal{L} &= \phi(\hat{y}_{cls},y)+\phi(Avg(\hat{y}),y),
    \end{split}
\end{equation}
where $y$ is the ground-truth of $x_{cls}$ and $x_{patch}$; $m$ denotes the transformer model; $\phi$ is the classification metric function, usually realized by the cross-entropy loss.
During inference, the final average pooled features are used for classification and the CLS token is only used for token selection. 

%\paragraph{Complexity Analysis}
%\begin{equation*}
%    \begin{split}
%        \mathcal{O}(SA) &= 4HWC^2 + 2(HW)^2C, \\
%        \mathcal{O}(Ours) &= xxx.
%    \end{split}
%\end{equation*}

%% file: Figs/fig_method.tex
\begin{figure}
    \centering
    \includegraphics[width=0.49\textwidth]{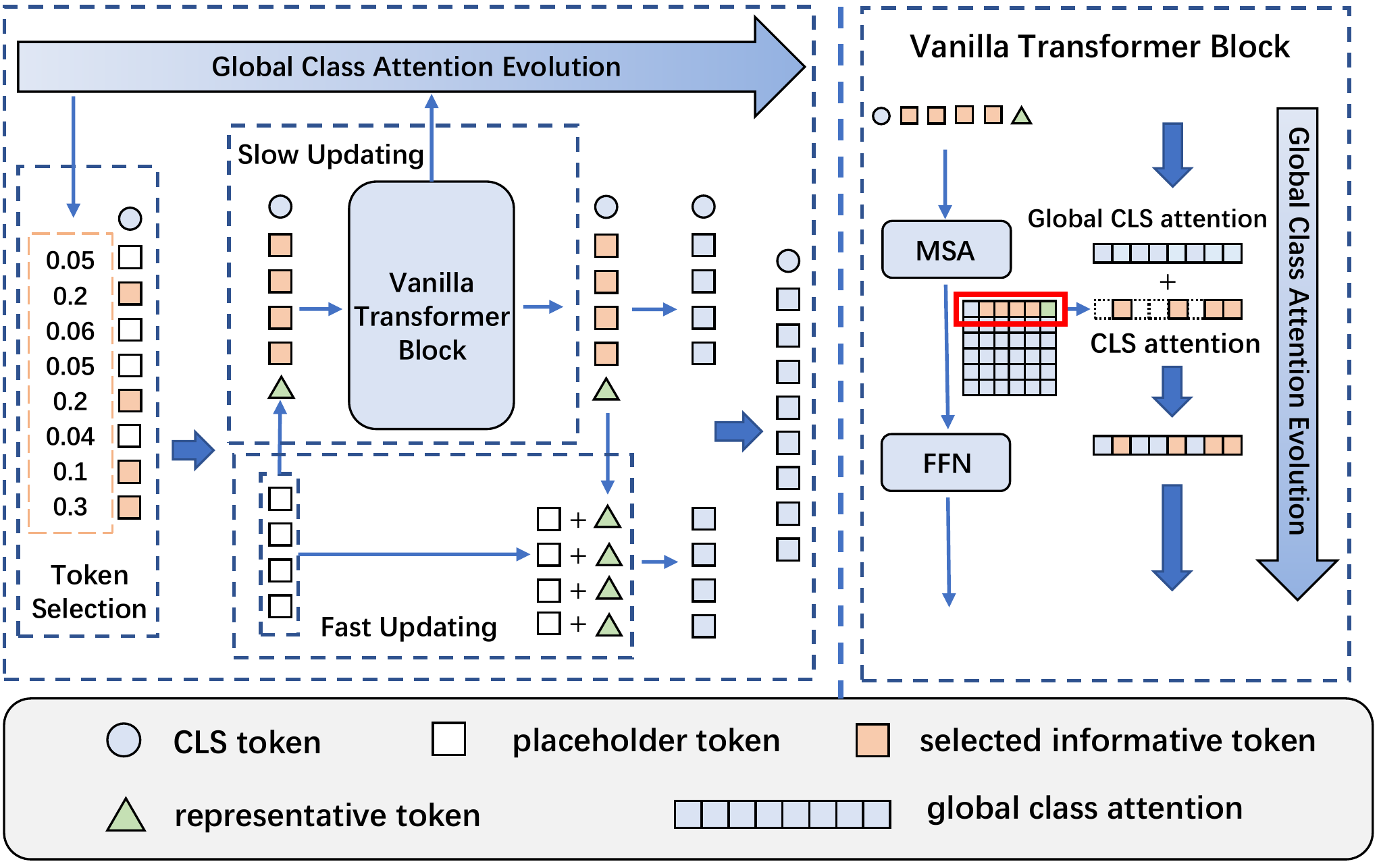} % Reduce the figure size so that it is slightly narrower than the column.
  \caption{%The overall diagram of the proposed slow-fast token evolution (Evo-ViT) method.
  The overall diagram of the proposed method.}
    \label{fig:method}
    \vspace{-4mm}
\end{figure}

%% file: Figs/fig_cka.tex
\begin{figure}[t]
    \centering  %图片全局居中
    \subfigure[Inter-layer]{
    \label{fig:cka_score}
    \includegraphics[width=0.43\linewidth]{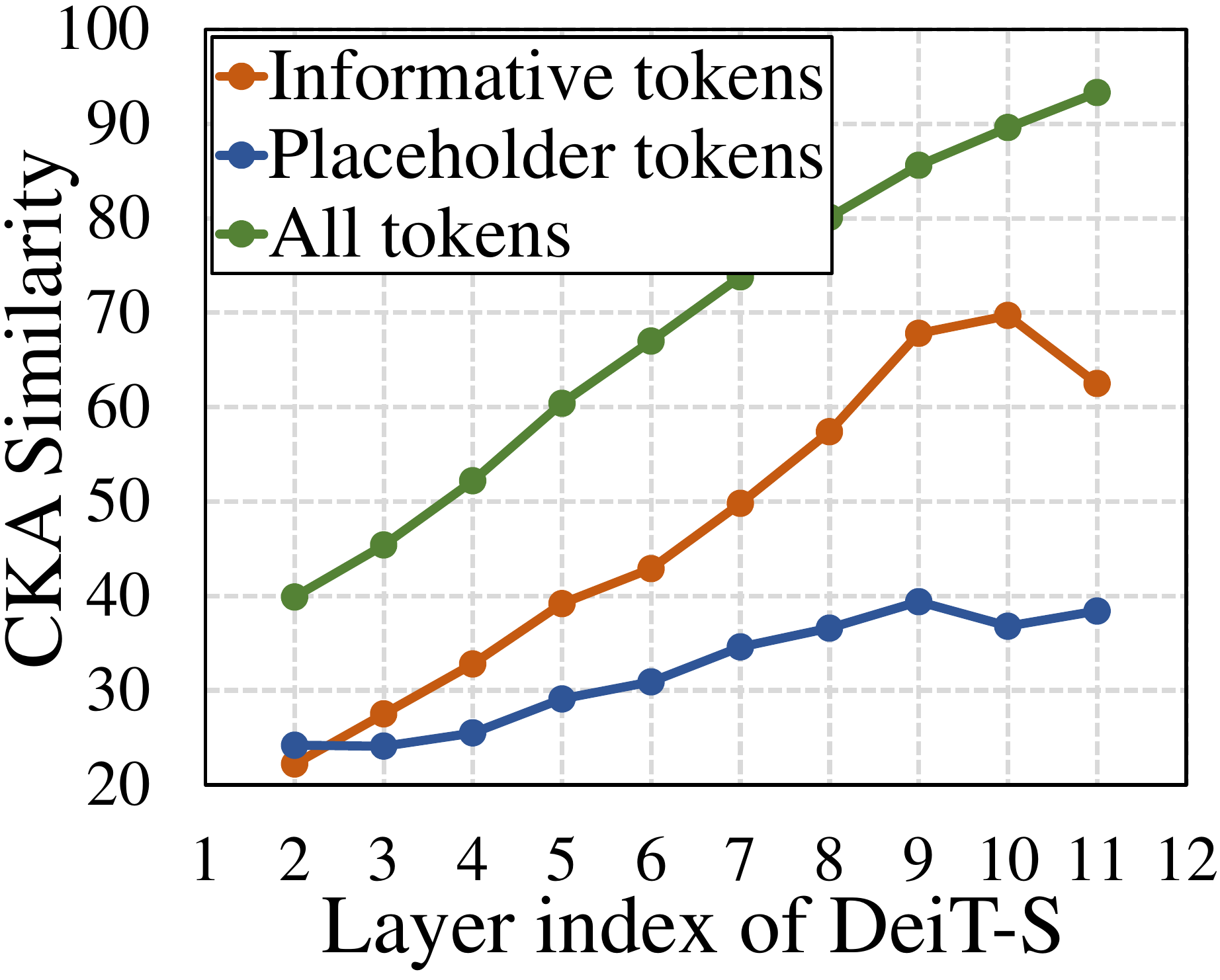}}
    \subfigure[Intra-layer]{
    \label{fig:token_correlation}
    \includegraphics[width=0.53\linewidth]{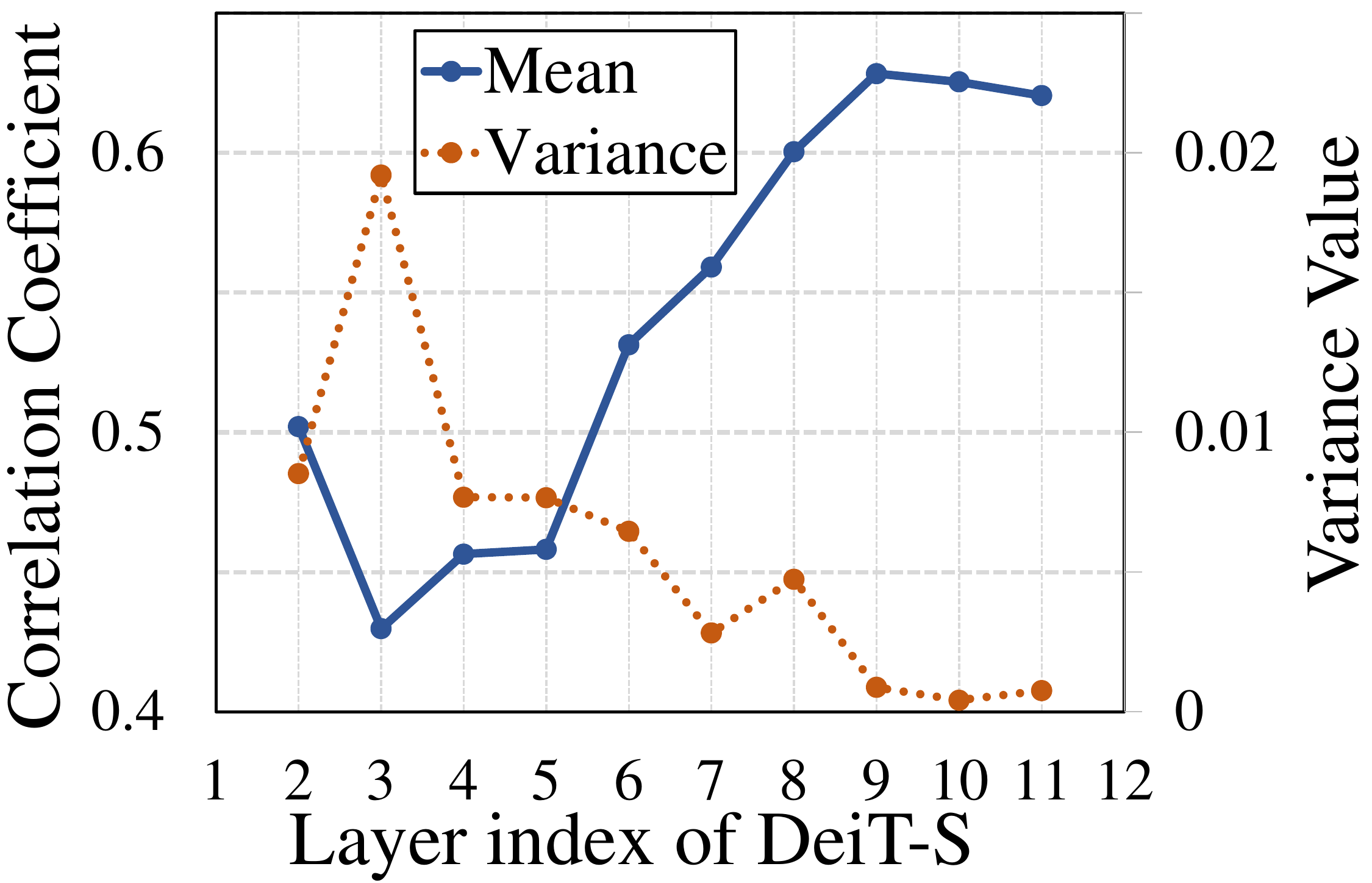}}
    \caption{Two folds that illustrate the difficulty of pruning the shallow layers. (a) The CKA similarity between the final CLS token and token features in each layer. (b) The Pearson correlation coefficient of the token features in each layer.}
    \label{fig:cor_analysis}
    \vspace{-5mm}
\end{figure}

%% file: Sections/Experiments_v3.tex
\section{Experiments}
\label{sec:experiments}

\subsection{Setup}
In this section, we demonstrate the superiority of the proposed Evo-ViT approach through extensive experiments on the ImageNet-1k~\cite{imagenet} classification dataset. To demonstrate the generalization of our method, we conduct experiments on vision transformers of both flat and deep-narrow structures, \emph{i.e.}, DeiT~\cite{DeiT} and LeViT~\cite{LeViT}. Following ~\cite{LeViT}, we train LeViT with distillation and without batch normalization fusion.
We apply the position embedding in~\cite{PVT} to LeViT for better efficiency.
For overall comparisons with the state-of-the-art methods~\cite{DynamicViT,PatchSlimming,SViTE,IA-RED2}, we conduct the token selection and slow-fast token updating from the fifth layer of DeiT and 
the third layer (excluding the convolution layers) of LeViT, respectively. The selection ratios of informative tokens in all selected layers of both DeiT and LeViT are set to 0.5.
The global CLS attention trade-off $\alpha$ in Eqn.~\ref{eq:global_attn} are set to 0.5 for all layers.
For fair comparisons, all the models are trained for 300 epochs.

\input{Tabs/comparision_prune_method}

\subsection{Main Results}
\noindent\textbf{Comparisons with existing pruning methods.}
In Table~\ref{tab:prune_compare}, we compare our method with existing token pruning methods~\cite{DynamicViT,IA-RED2,PatchSlimming,SViTE}. Since token pruning methods are unable to recover the 2D structure and are usually designed for transformers with flat structures, we comprehensively conduct the comparisons based on DeiT~\cite{DeiT} on ImageNet dataset. We report the top-1 accuracy and throughput for performance evaluation. The throughput is measured on a single NVIDIA V100 GPU with batch size fixed to 256, which is the same as the setting of DeiT. Results indicate that our method outperforms previous token pruning methods on both accuracy and efficiency. Our method accelerates the inference throughput by over 60$\%$ with negligible accuracy drop (-0.4$\%$) on DeiT-S.

\noindent\textbf{Comparisons with state-of-the-art ViTs.}
Owing to the preserved placeholder tokens, our method guarantees the spatial structure that is indispensable for most existing modern ViT networks. Thus, we further apply our method to the state-of-the-art efficient transformer LeViT~\cite{LeViT}, which presents a deep-narrow architecture. As shown in Table~\ref{tab:main_compare}, our method can further accelerate the deep-narrow transformer such as LeViT. We have observed larger accuracy degradation of our method on LeViT than on DeiT. The reason lies that the deeper layers of LeViT have few tokens and therefore have less redundancy due to the shrinking pyramid structure. With dense input, such as the image resolution of 384×384, our method accelerates LeViT with less accuracy degradation and more acceleration ratio, which indicates the effectiveness of our method on dense input.

\input{Tabs/main_exp}
\subsection{Ablation Analysis}
\textbf{Effectiveness of each module.}
To evaluate the effectiveness of each sub-method, we add the following improvements step by step in Tab.~\ref{tab:method_ablation_deit} on transformers of both flat and deep-narrow structures, namely DeiT and LeViT:
\textit{i) Naive selection}. Simply drop the placeholder tokens based on the original class attention in each layer;
\textit{ii) Structure preservation}. Preserve the placeholder tokens but not fast update them;
\textit{iii) Global attention}. Utilize the proposed global class attention instead of vanilla class attention for token selection;
\textit{iv) Fast updating}. Augment the preserved placeholder tokens with fast updating;
\textit{v) Layer-to-stage}. Apply the proposed layer-to-stage training strategy to further accelerate inference.

Results on DeiT indicate that our structure preserving strategy further improves the selection performance due to its capacity of preserving complete information flow. The evolved global class attention enhances the consistency of token selection across layers and achieves better performance. The fast updating strategy has less effect on DeiT than on LeViT. We claim that the performance of DeiT turns to be saturated based on structure preservation and global class attention, while LeViT still has some space for improvement. LeViT exploits spatial pooling for token reduction, which makes unstructured token reduction in each stage more difficult. By using the fast updating strategy, it is possible to collect some extra cues from  placeholder tokens for accurate and augmented feature representations.
We also evaluate the layer-to-stage strategy. Results indicate that it further accelerates inference while maintaining accuracy. 
\input{Tabs/method_ablation_deit}

% \paragraph{Hyper parameter analysis.}
% We further investigate the hyper parameters of our method on DeiT-T, namely, keeping ratio, starting layer index, global attention trade-off, and starting epoch. We initialize these hyper parameters as described in Setup. During ablation analysis, only the object parameter is changed and the others remain fixed. The layer-to-stage training strategy is disabled in the following ablation experiments.

% \textit{Keeping ratio} determines how many tokens are kept for slow updating in each layer. For precision, we set all layers with the same keeping ratio and investigate the trade-off between accuracy and inference throughput in Fig.~\ref{fig:pruning_ratio_ablation}. Results show that the accuracy turns to be saturated when the keeping ratio is larger than 0.5. Another interesting finding is that the accelerated model with 0.9 keeping ratio outperforms the full baseline by 0.2$\%$ (72.2 to 72.4), which is consistent with the conclusion in \cite{SViTE} that properly dropping several uninformative tokens can serve as regularization for vision transformers.

% \input{Figs/fig_ablation2}
\input{Tabs/subsample_strategy_comparision}
\input{Figs/fig_prunevis}

% \textit{Starting layer index} denotes which layer to start token selection and slow-fast updating. Fig.~\ref{fig:start_layer_ablation} indicates that the accuracy turns to be stable when we start from the fifth layer. We find a significant accuracy degradation as the starting layer becomes shallow, especially for the first three layers. We claim the reason lies that the features in these layers are still with large variation and not stable as shown in Fig.~\ref{fig:cka_score} and Fig.~\ref{fig:token_correlation}.

% \textit{Global attention trade-off} in Eqn.~\ref{eq:global_attn} controls the dependence on previous layer information when conduct token selection in each layer. Larger trade-off leads to stronger dependence on previous information. Fig.~\ref{fig:global_attn_tradeoff_ablation} illustrates that it is the best to equally consider the previous and current information.

% \textit{Starting epoch} denotes which epoch to start our token selection and updating strategy. As shown in Fig.~\ref{fig:start_epoch}, the accuracy drop sharply when we start from the last 100 epoch. We claim that the dynamic token selection requires enough training epochs to learn the refined features. For precision and training efficiency, we start our token selection and evolution strategy from the very beginning of training.

\noindent\textbf{Different Token Selection Strategy.}
We compare our global-attention-based token selection strategy with several common token selection strategies and sub-sampling methods in Tab.~\ref{tab:subsample_strategy_comparision} to evaluate the effectiveness of our method. All token selection strategies are conducted under our structure preserving strategy without layer-to-stage training schedule.
The token selection strategies include: randomly selecting the informative tokens (\textit{random selection}); Utilizing the class attention of the last layer for selection in all layers via twice inference (\textit{last class attention});  taking the column mean of the attention matrix as the score of each token as proposed in~\cite{kim2021learned} (\textit{attention column mean}).

Results in Tab.~\ref{tab:subsample_strategy_comparision} indicate that our evolved global class attention outperforms the other selection strategies and common sub-sampling methods on both accuracy and efficiency. We have observed obvious performance degradation with last class attention, although the attention in deeper layers is more focused on objects in Fig.~\ref{fig:fig1}. A possible reason is that the networks require some background information to assist classification, while restricting all layers to only focus on objects during the entire training process leads to underfitting on the background features.

\input{Figs/different_ratio_ablation}

\noindent\textbf{Visualization.}
We visualize the token selection in Fig.~\ref{fig:prune_vis} to demonstrate performance of our method during both training and testing stages. The visualized models in the left and middle three columns are trained without the layer-to-stage training strategy.
The left three columns demonstrate results on different layers of a well-trained DeiT-T model. 
Results show that our token selection method mainly focuses on objects instead of backgrounds, thereby indicating that our method can effectively discriminate the informative tokens from placeholder tokens. 
The selection results tend to be consistent across layers, which proves the feasibility of our layer-to-stage training strategy.
Another interesting finding is that some missed tokens in the shallow layers are retrieved in the deep layers owing to our structure preserving strategy. 
Take the baseball images as an example, tokens of the bat are gradually picked up as the layer goes deeper. 
This phenomenon is more obvious under our layer-to-stage training strategy in the right three columns. 
We also investigate how the token selection evolves during the training stage in the middle three columns. Results demonstrate that some informative tokens, such as the fish tail, are determined as placeholder tokens at the early epochs. With more training epochs, our method gradually turns to be stable for discriminative token selection.

\noindent\textbf{Consistent keeping ratio.} 
We set different keeping ratio of tokens in each layer to investigate the best acceleration architecture of Evo-ViT. The keeping ratio determines how many tokens are kept as informative tokens. Previous token pruning works~\cite{DynamicViT,PatchSlimming} present a gradual shrinking architecture, in which more tokens are recognized as placeholder tokens in deeper layers. They are restricted in this type of architecture due to direct pruning. Our method allows more flexible token selection owing to the structure preserving slow-fast token evolution. As shown in Fig.~\ref{fig:abl_diff_pr}, we maintain the sum of the number of placeholder tokens in all layers and adjust the keeping ratio in each layer. Results demonstrate that the best performance is reached with a consistent keeping ratio across all layers.
We explain the reason as follows. 
In the above visualization, we find that the token selection results tend to be consistent across layers, indicating that the transformer tends to augment informative tokens with computing resource as much as possible. 
In Fig.~\ref{fig:abl_diff_pr}, at most 50$\%$ tokens are passed through the full transformer network when the keeping ratios in all layers are set to $0.5$, thereby augmenting the most number of informative tokens with the best computing resource, namely, the full transformer network.

%% file: Tabs/comparision_prune_method.tex
\begin{table}
    \centering
    \caption{Comparison with existing token pruning methods on DeiT. The image resolution is $224 \times 224$ unless specified. $^{*}$ denotes that the image resolution is $384 \times 384$.}
    \label{tab:prune_compare}
    \small{
    \setlength{\tabcolsep}{1.5mm}{
    \begin{tabular}{l|ccc}
    \toprule
    \multirow{2}{*}{Method} & Top-1 Acc. & \multicolumn{2}{c}{Throughput} \\
     & (\%) & (img/s) & (\%) \\
    \midrule
    \multicolumn{4}{c}{DeiT-T} \\
    \midrule
    Baseline~\cite{DeiT} & 72.2 & 2536 & -  \\
    PS-ViT~\cite{PatchSlimming}     &      72.0       &        3563       &       40.5      \\
    DynamicViT~\cite{DynamicViT}    &      71.2       &        3890       &       53.4        \\
    SViTE~\cite{SViTE}              &      70.1       &          2836     &       11.8      \\
    \textbf{Evo-ViT (ours)}         &    72.0         & \textbf{4027}     &  \textbf{58.8}  \\
    \midrule
    \multicolumn{4}{c}{DeiT-S} \\
    \midrule
    Baseline~\cite{DeiT}            &      79.8       &                940        &       -         \\
    % Baseline$^{*}$~\cite{DeiT}            &              &             258            &       -         \\
    PS-ViT~\cite{PatchSlimming}     &      79.4       &                1308       &       43.6      \\
    DynamicViT~\cite{DynamicViT}    &      79.3       &                 1479      &       57.3        \\
    SViTE~\cite{SViTE}              &      79.2       &                1215       &       29.3      \\
    IA-RED$^2$~\cite{IA-RED2}       &      79.1       &                 1360      &       44.7      \\
    \textbf{Evo-ViT (ours) }        &      79.4   &        \textbf{1510}      &  \textbf{60.6}      \\
    % \textbf{Evo-ViT$^{*}$ (ours) }  &         &      421        &    63.1    \\
    \midrule
    \multicolumn{4}{c}{DeiT-B} \\
    \midrule
    Baseline~\cite{DeiT}            &      81.8       &                 299        &       -         \\
    Baseline$^{*}$~\cite{DeiT}      &      82.8       &                87        &       -         \\
    PS-ViT~\cite{PatchSlimming}     &      81.5       &                445        &       48.8      \\
    DynamicViT~\cite{DynamicViT}    &      80.8       &                454        &       51.8      \\
    SViTE~\cite{SViTE}              &      82.2   &                 421        &       40.8      \\
    IA-RED$^2$~\cite{IA-RED2}       &      80.9       &                453        &       42.9      \\
    IA-RED$^{2*}$~\cite{IA-RED2}    &      81.9       &                129        &       51.5      \\
    \textbf{Evo-ViT (ours) }        &      81.3       &      \textbf{462}        &       \textbf{54.5} \\
    \textbf{Evo-ViT$^{*}$ (ours)}   &      82.0       &                \textbf{139}        &       \textbf{59.8}      \\
    \bottomrule
    \end{tabular}
    }
    }
    \vspace{-3mm}
\end{table}

%% file: Tabs/main_exp.tex
\begin{table}[t]
    \center
    \caption{Comparison with state-of-the-art vision transformers. The input image resolution is $224 \times 224$ unless specified. $^{*}$ denotes that the image resolution is $384 \times 384$.}
    \label{tab:main_compare}
    \small{
    \setlength{\tabcolsep}{2.8mm}{
    \begin{tabular}{l|cccc}
    \toprule
    \multirow{2}{*}{Model}         & Param & Throughput & Top-1 Acc. \\
             & (M) & (img/s) & (\%) \\
    \midrule
    % DeiT-T        &     5.9           &     2536   &      72.2     \\
    LeViT-128S    &     7.8           &     8755   &     74.5      \\
    LeViT-128     &     9.2           &     6109   &     76.2      \\
    LeViT-192     &     10.9          &     4705   &     78.4      \\
    PVTv2-B1      &     14.0          &     1225   &     78.7      \\
    CoaT-Lite Tiny&     5.7           &     1083   &      76.6         \\
    % T2T-ViT-7    &     4.2           &      2012      &      71.2         \\ \midrule
    PiT-Ti    &     4.9           &      3030      &      73.0         \\ \midrule
    % \textbf{Evo-DeiT-Ti}    &     5.9           &     3978   &  72.0         \\ 
    \textbf{Evo-LeViT-128S} &     7.8           &     10135  &     73.0      \\
    \textbf{Evo-LeViT-128}  &     9.2           &     8323   &      74.4     \\
    \textbf{Evo-LeViT-192}  &     11.0          &    6148    &      76.8     \\ \midrule\midrule
    % DeiT-S        &     22.5          &    940     &      79.8     \\
    LeViT-256     &     18.9          &    3357    &      80.1     \\
    LeViT-256$^{*}$&     19.0          &   906     &        81.8   \\
    PVTv2-B2      &     25.4          &   687     &      82.0         \\
    % T2T-ViT-14    &     21.4           &   793      &      80.6         \\
    PiT-S    &     23.5           &   1266      &      80.9         \\
    Swin-T      &    29.4          &       755     &     81.3      \\
    CoaT-Lite Small&     20.0         &     550      &      81.9         \\\midrule
    % \textbf{Evo-DeiT-S}     &     22.5          &    1510    &    79.4       \\
    \textbf{Evo-LeViT-256}  &     19.0          &    4277    &     78.8      \\ 
    \textbf{Evo-LeViT-256$^{*}$}  &     19.2          &    1285    &  81.1     \\ \midrule\midrule
    % DeiT-B        &     86.2          &    317     &     81.8      \\
    % DeiT-B$^{*}$  &     86.2          &    87     &      82.8     \\ 
    LeViT-384     &     39.1          &     1838   &     81.6      \\
    LeViT-384$^{*}$ &     39.2          &   523     &     82.8      \\
    PVTv2-B3      &    45.2          &      457     &     83.2      \\
    % Swin-S      &    50.3          &      436     &     83.0      \\
    % T2T-ViT-19    &     39.0           &    486       &      81.2         \\ \midrule
    PiT-B    &     73.8           &    348       &      82.0         \\ \midrule
    % \textbf{Evo-DeiT-B}     &     86.2          &    455     &     81.3      \\
    \textbf{Evo-LeViT-384}  &     39.3          &    2412    &     80.7      \\
    \textbf{Evo-LeViT-384$^{*}$}  &     39.6          &    712    &   82.2        \\
    \bottomrule
    \end{tabular}
    }
    }
    \vspace{-3mm}
\end{table}

%% file: Tabs/method_ablation_deit.tex
\begin{table}[t]
\centering
\caption{Method ablation on DeiT and LeViT.}
\label{tab:method_ablation_deit}
\setlength{\tabcolsep}{1mm}
\small{
\begin{tabular}{l|cc|cc}
\toprule
\multirow{3}{*}{Strategy} & \multicolumn{2}{c}{DeiT-T}           & \multicolumn{2}{|c}{LeViT 128S} \\
\cline{2-5}
 & Acc. & Throughput & Acc. & Throughput \\
 & (\%) & (img/s) & (\%) & (img/s) \\
\midrule
baseline                 & 72.2            & 2536               & 74.5            & 8755               \\
+ naive selection        & 70.8            & 3824               & -               & -                  \\
+ structure preservation & 71.6            & 3802               & 72.1            & 9892               \\
+ global attention       & 72.0            & 3730               & 72.5            & 9452               \\
+ fast updating          & 72.0            & 3610               & 73.0            & 9360               \\
+ layer-to-stage         & 72.0            & 4027               & 73.0            & 10135               \\
\bottomrule
\end{tabular}
}
\vspace{-2mm}
\end{table}

%% file: Tabs/subsample_strategy_comparision.tex
\begin{table}[t]
\centering
\caption{Different token selection strategies on DeiT-T. 
% To align the throughput, 
We conduct all sub-sampling methods at the seventh layer and conduct token selection strategies from the fifth layer.}
\label{tab:subsample_strategy_comparision}
\small{
\setlength{\tabcolsep}{2.4mm}{
\begin{tabular}{lcc}
\toprule
Method                 & Acc. (\%) & Throughput (img/s)  \\
\midrule
average pooling        &      69.5       &        3703         \\
max pooling            &      69.8       &        3698         \\
convolution            &      70.2       &        3688         \\ \midrule
random selection       &      66.4       &        3760         \\
last class attention   &      69.7       &        1694         \\
attention column mean  &      71.2       &        3596         \\       
global class attention &      72.0       &        3730         \\
\bottomrule
\end{tabular}
}
}
\vspace{-3mm}
\end{table}

%% file: Figs/fig_prunevis.tex
\begin{figure*}
    \centering
    \includegraphics[width=\textwidth]{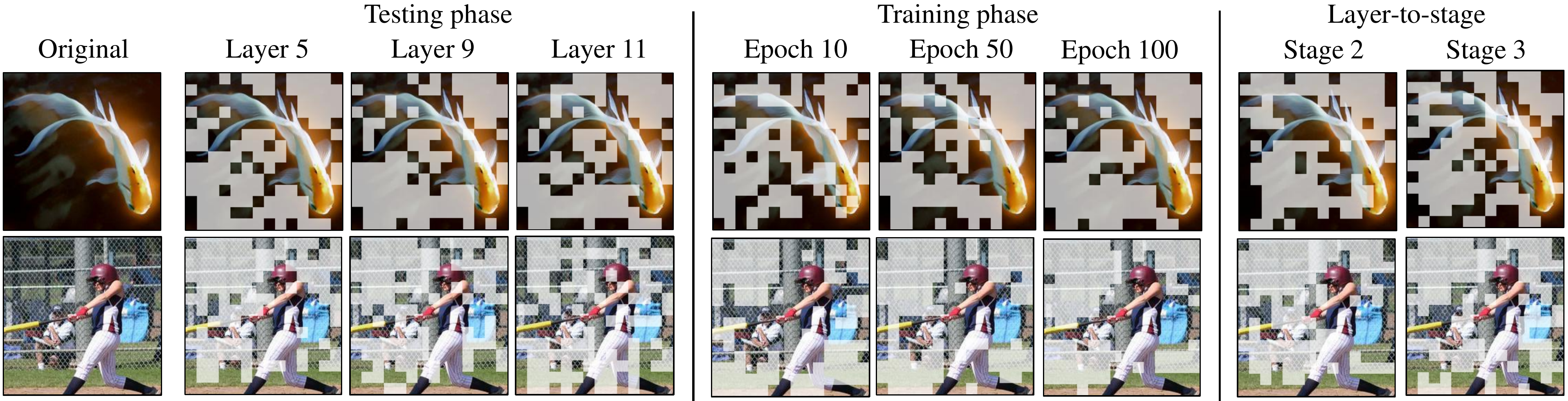} % Reduce the figure size so that it is slightly narrower than the column.
    \caption{Token selection results on DeiT-T. The left, middle, and right three columns denote the selection results on a well-trained Evo-ViT, the fifth layer at different training epochs, and Evo-ViT with the proposed layer-to-stage strategy, respectively.
    % The left three columns demonstrate results on different layers of a well-trained model. The middle three columns demonstrate results on the fifth layer at different training epochs.
    }
    \label{fig:prune_vis}
    \vspace{-3mm}
\end{figure*}

%% file: Figs/different_ratio_ablation.tex
\begin{figure}[t]
    \centering
    \includegraphics[width=0.48\textwidth]{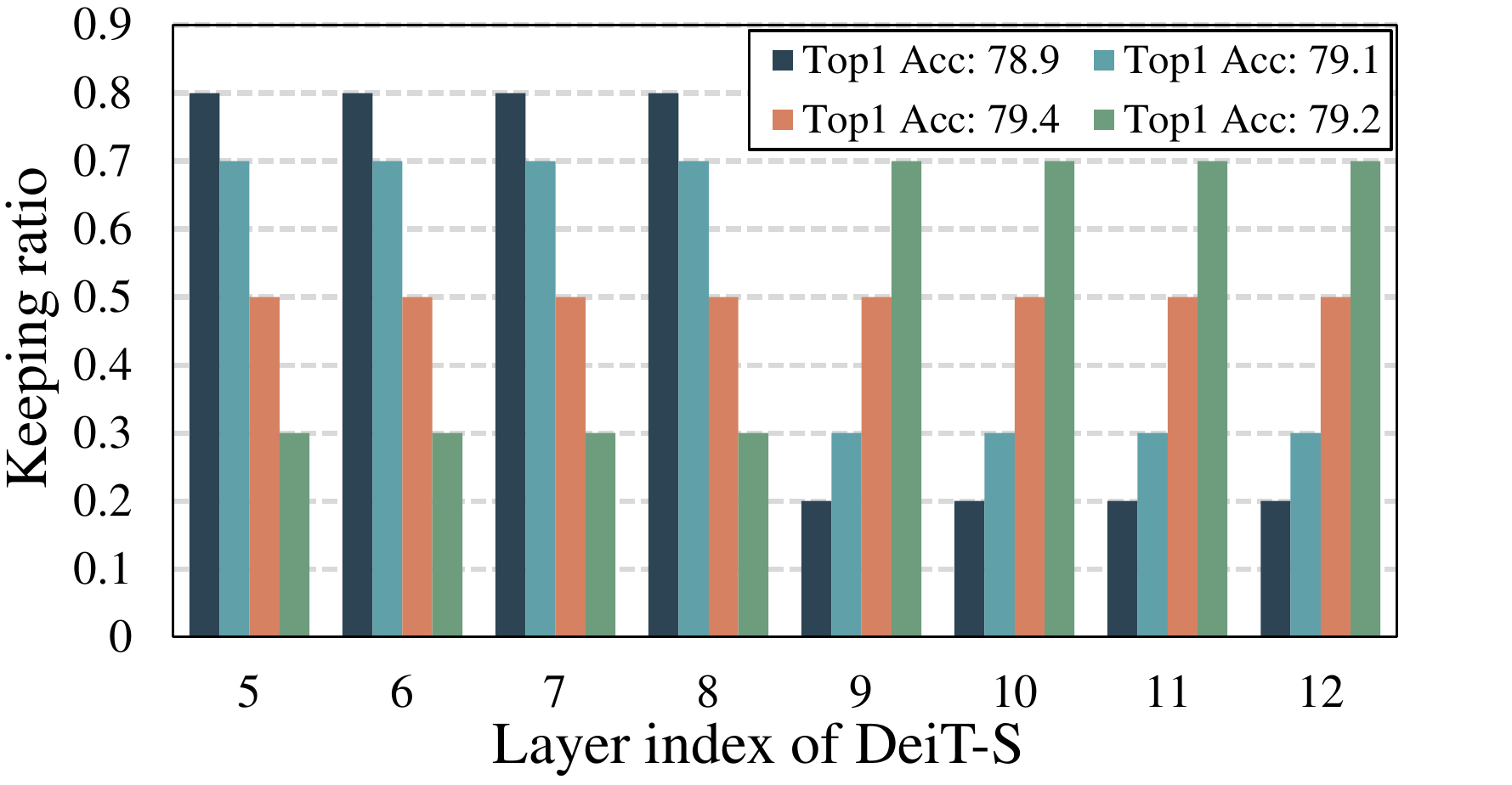} % Reduce the figure size so that it is slightly narrower than the column.
    \caption{Different architecture of the accelerated DeiT-S via our method. We start our token selection from the fifth layer.}
    \label{fig:abl_diff_pr}
    \vspace{-5mm}
\end{figure}

%% file: Sections/Conclusions.tex
\section{Conclusions}
\label{sec:conclusions}
In this work, we investigate the efficiency of vision transformers by developing a self-motivated slow-fast token evolution (Evo-ViT) method. We propose the structure preserving token selection and slow-fast updating strategies to solve the limitation of token pruning on modern structured compressed transformers and scatch training. Extensive experiments on two popular ViT architectures, \emph{i.e.}, DeiT and LeViT, indicate that our Evo-ViT approach is able to accelerate various transformers significantly while maintaining comparable classification performance.

% outperforming existing token pruning methods on both accuracy and efficiency.
As for future work, an interesting and worthwhile direction is to extend our method to downstream tasks, such as object detection and instance segmentation. 
% For each instance, the informative tokens and placeholder tokens are determined by the evolved global class attention of the transformer. By preserving placeholder tokens and updating them in a fast way, both the complete information flow and spatial structure are preserved for training stability and generalization to transformers of flat and deep-narrow structures. 
% Experiments on DeiT and LeViT indicate that the proposed  Evo-ViT  significantly accelerates transformers while maintaining comparable classification performance.